\documentclass[11pt,a4paper]{article}
\usepackage[hyperref]{naaclhlt2019}
\usepackage{times}
\usepackage{latexsym}
\usepackage{amsmath}
\usepackage{subcaption}
\usepackage{amsmath}
\usepackage{multirow}
\usepackage{booktabs}
\usepackage{forest}
\usepackage{float}
\usepackage{times}
\usepackage{multirow}
\usepackage{booktabs}
\usepackage{tabularx}
\usepackage{scalefnt}
\usepackage{graphicx}
\usepackage{tikz}
\usepackage{verbatim}
\usepackage{graphicx}

\usepackage{url}

\aclfinalcopy 
\def\aclpaperid{65} 

\usepackage[]{todonotes}  

\title{Identifying and Reducing Gender Bias in Word-Level Language Models}

\author{
Shikha Bordia$^{1}$\\
{\tt sb6416@nyu.edu} 
\And
Samuel R. ~Bowman$^{1,2,3}$\\
{\tt bowman@nyu.edu}
\AND
$^{1}$\normalfont Dept. of Computer Science\\New York University\\251 Mercer St\\New York, NY 10012 \And
$^{2}$\normalfont Center for Data Science\\New York University\\60 Fifth Avenue\\New York, NY 10011\And
$^{3}$\normalfont Dept. of Linguistics\\New York University\\10 Washington Place\\New York, NY 10003
  }

\begin{document}
\maketitle
\begin{abstract}
 
Many text corpora exhibit socially problematic biases, which can be propagated  or  amplified  in  the  models trained on such data. For example, \textit{doctor} cooccurs more frequently with male pronouns than female pronouns. In this study we (i) propose a metric  to measure gender bias; (ii) measure bias in a text corpus and the text generated from a recurrent  neural  network language model trained on the text corpus; (iii) propose a regularization loss term for the language model that minimizes the projection of encoder-trained embeddings onto an embedding subspace that encodes gender; (iv) finally, evaluate efficacy of our proposed method on reducing gender bias. We find this regularization method to be effective in reducing gender bias up to  an  optimal weight assigned to the loss term, beyond  which the model becomes unstable as the perplexity increases. We replicate this study on three training corpora---Penn Treebank,  WikiText-2, and CNN/Daily Mail---resulting in similar conclusions.

\end{abstract}

\section{Introduction}

Dealing with discriminatory bias in training data is a major issue concerning the mainstream implementation of machine learning. Existing biases in  data can be amplified by models and the resulting output consumed by the public can influence them, encourage and reinforce harmful stereotypes, or distort the truth. Automated systems that depend on these models can take problematic actions based on biased profiling of individuals. The National Institute for Standards and Technology (NIST) evaluated several facial recognition algorithms and found that they are systematically biased based on gender \cite{ngan2015face}. Algorithms performed worse on faces labeled as female than those labeled as male. 


Models automating resume screening have also proved to have a heavy gender bias  favoring male candidates \citep{lambrecht2018algorithmic}. Such data and algorithmic biases have become a growing concern. Evaluation and mitigation of  biases in  data and models that use the data has been a growing field of research in recent years.

One natural language understanding task vulnerable to gender bias is language modeling. The task of language modeling has a number of practical applications, such as word prediction used in onscreen keyboards. 
If possible, we would like to identify the bias in the data used to train these models and reduce its effect on model behavior.

Towards this pursuit, we aim to evaluate the effect of gender bias on word-level language models that are trained on a text corpus.
Our contributions in this work include: (i) an analysis of the gender bias exhibited by publicly available datasets used in building state-of-the-art language models; (ii) an analysis of the effect of this bias on recurrent neural networks (RNNs) based word-level language models; (iii) a method for reducing bias learned in these models; and (iv) an analysis of the results of our method.

\begin{figure*}[hbt!]
    \centering
    \includegraphics[width=14cm]{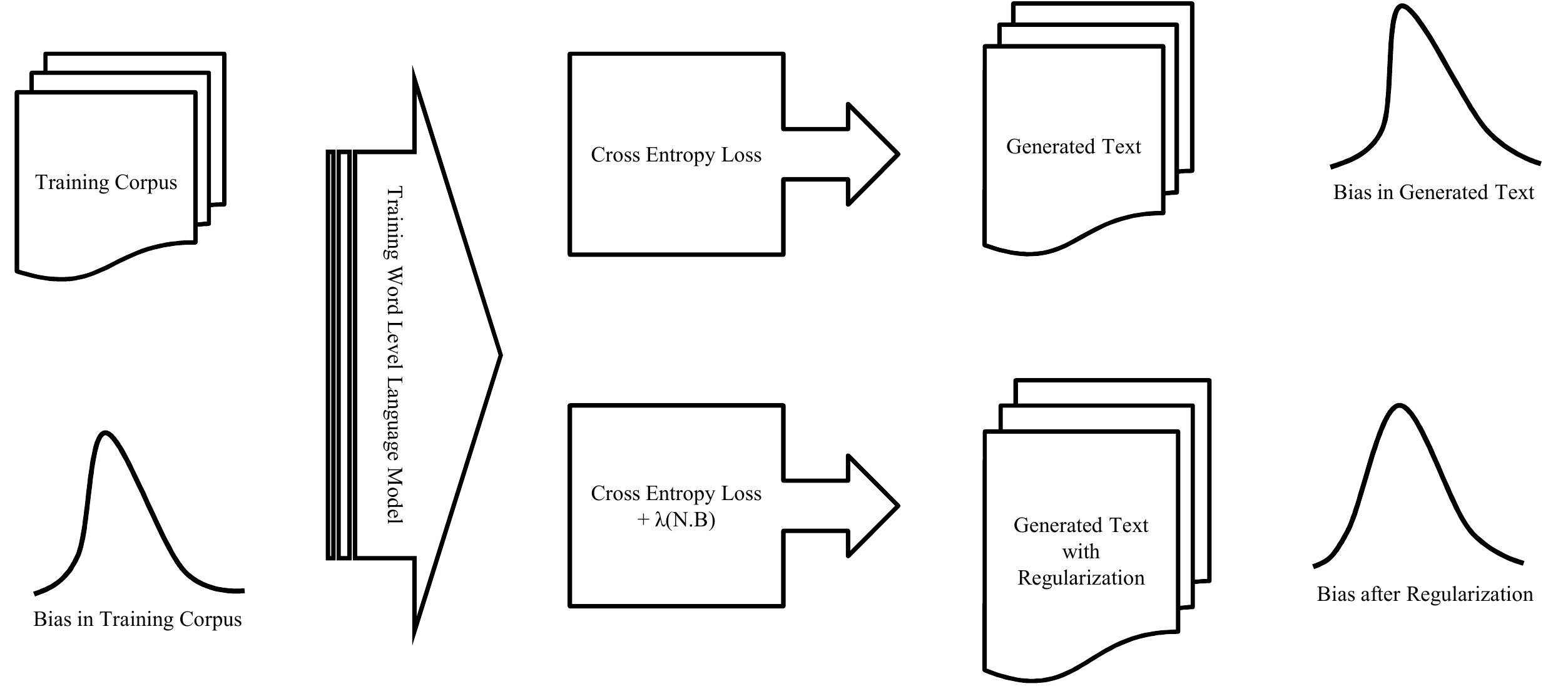}
    \caption{Word level language model is a three layer LSTM model. $\lambda$ controls the importance of minimizing bias in the embedding matrix.  }
    \label{fig:flowchart}
\end{figure*}

\section{Related Work}

A number of methods have been proposed for evaluating and addressing biases that exist in datasets and the models that use them. \newcite{recasens2013linguistic} studies the neutral point of view (NPOV) edit tags in the Wikipedia edit histories to understand linguistic realization of bias. According to their study, bias can be broadly categorized into two classes: framing and epistemological. While the framing bias is more explicit, the epistemological bias is implicit and subtle. Framing bias occurs when subjective or one-sided words are used. For example, in the sentence---\textit{``Usually, smaller cottage-style houses have been demolished to make way for these McMansions.''}, the word \textit{McMansions} has a negative connotation towards large and pretentious \textit{houses}. Epistemological biases are entailed, asserted or hedged in the text. For example, in the sentence---\textit{``Kuypers claimed that the mainstream press in America tends to favor liberal viewpoints,''} the word \textit{claimed} has a doubtful effect on Kuypers statement as opposed to \textit{stated} in the sentence---\textit{``Kuypers stated that the mainstream press in America tends to favor liberal viewpoints.''} It may be possible to capture both of these kinds of biases through the distributions of co-occurrences. In this paper, we deal with identifying and reducing gender bias based on words co-occurring in a context window. 


\newcite{bolukbasi2016man}  propose an approach to investigate gender bias present in popular word embeddings, such as word2vec  \citep{mikolov2013distributed}. They construct a gender subspace  using a set of binary gender pairs. For words that are not explicitly gendered, the component of the word embeddings that project onto this subspace can be removed to debias the embeddings in the gender direction. They also propose a softer variation that balances reconstruction of the original embeddings while minimizing the part of the embeddings that project onto the gender subspace. We use the softer variation to debias the embeddings while training our language model.

\newcite{Zhao2017men} look at gender bias in the context of using structured prediction for visual object classification and semantic role labeling. They observe gender bias in the training examples and that their model amplifies the bias in its predictions. 
They impose constraints on the optimization to reduce bias amplification while incurring minimal degradation in their model's performance.

Word embeddings can capture the stereotypical bias in human generated text leading to biases in NLP Applications.  \newcite{caliskan2016} conduct \textit{Word Embedding Association Test} (WEAT). It is based on the hypothesis that word embeddings closer together in high dimensional space are semantically closer. They find strong evidence  of social biases in pretrained word embeddings.

\newcite{rudinger-EtAl:2018:N18} introduce \textit{Winogender schemas}\footnote{It is Winograd Schema-style coreference dataset consisting of pair of sentences that differ only by a gender pronoun} and  evaluate three coreference resolution systems---rule-based, statistical and neural systems. They find that these systems' predictions  strongly prefer one gender over the other for occupations. 

\newcite{Joel2019} study the impact of gender debiasing techniques by \citet{bolukbasi2016man} and \citet{zhao2018b} in machine translation. They find these methods to be effective, and even a noted BLEU score improvement for the debiased model. Our work is closely related but while they use debiased pretrained embeddings, we train the word embeddings from scratch and debias them while the language model is trained. 

\newcite{Chandler2019} extend WEAT to state-of-the-art sentence encoders: the Sentence Encoder Association Test (SEAT). They show that these tests can provide an evidence for presence of bias. However, the cosine similarity between sentences can be an inadequate measure of text similarity in sentences. In this paper, we attempt to minimize the cosine similarity between word embeddings and gender direction. 


\newcite{Gonen2019} conduct experiments using the debiasing techniques proposed by \citet{bolukbasi2016man} and \citet{zhao2018b}. They show that bias removal techniques based on gender direction  are inefficient in removing all aspects of bias. In a high dimensional space, spatial distribution of the gender neutral word embeddings remain almost same after debiasing. This enables a gender-neutral classifier to still pick up the cues that encode other semantic aspects of bias. We use softer variation of the debiasing method proposed by \citet{bolukbasi2016man} and attempt to measure the debiasing effect from the minimal changes in the embedding space.

\section{Methods}
We first examine the bias existing in the datasets through qualitative and quantitative analysis of trained embeddings and cooccurrence patterns. We then train an LSTM word-level language model on a dataset and measure the bias of the generated outputs. As shown in Figure \ref{fig:flowchart}, we then apply a regularization procedure that encourages the embeddings learned by the model to depend minimally on gender. We debias the input and the output embeddings individually as well as simultaneously.  
Finally, we assess the efficacy of the proposed method in reducing bias. 

We observe that when both input and output embeddings are debiased together, the perplexity of the model shoots up by a much larger number than the input or the output embeddings debiased individually. We report our results when only input embeddings are debiased.  This method, however, does not limit the model to capture other forms of bias being learned in other model parameters or output embeddings. 

The code implementing our methods can be found in our GitHub repository.\footnote{https://github.com/BordiaS/language-model-bias}

    


\subsection{Datasets and Text Preprocessing}

We compare the model on three datasets--Penn Treebank (PTB), WikiText-2 and CNN/Daily Mail. The first two have been used in language modeling for a long time. We include CNN/Daily Mail dataset in our experiments as it contains a more diverse range of topics. 

\paragraph{PTB} Penn Treebank comprises of articles ranging from scientific abstracts, computer manuals, etc. to news articles. 
In our experiments, we observe that PTB has a higher count of male words than female words. Following prior language modeling work, we use the Penn Treebank dataset \citep[PTB;][]{DBLP:journals/coling/MarcusSM94} preprocessed by \newcite{mikolov2010}.

 \paragraph{WikiText-2} WikiText-2 is twice the size of the PTB and is sourced from curated Wikipedia articles. It is more diverse  and therefore has a more balanced ratio of female to male gender words than PTB.  We use preprocessed WikiText-2 \citep[Wikitext-2;][]{merity2016pointer}. 

\paragraph{CNN/Daily Mail} This dataset is curated from a diverse range of news articles on topics like sports, health, business, lifestyle, travel etc. This dataset has an even more balanced ratio of female to male gender words and thus, relatively less biased than the above two. However, this does not mean that the use of pronouns is not biased. This dataset was released as part of a summarization dataset by \citet{nips15_hermann}, and contains 219,506 articles from the newspaper the Daily Mail. We sub-sample the sentences by a factor of 100 in order to make the dataset more manageable for experiments.

\subsection{Word-Level Language Model}

We use a three-layer LSTM word-level language model \citep[AWD-LSTM;][]{merity2017} with 1150 hidden units implemented in PyTorch.\footnote{https://github.com/salesforce/awd-lstm-lm} These models have an embedding size of 400 and a learning rate of 30. 

We use a batch size of 80 for Wikitext-2 and 40 for PTB. Both are trained for 750 epochs. The PTB baseline model achieves a perplexity of 62.56. For WikiText-2, the baseline model achieves a  perplexity of 67.67. 

For CNN/Daily Mail, we use a batch size of 80 and train it for 500 epochs. We do early stopping for this model. The hyperparameters are chosen through a systematic trial and error approach. The baseline model achieves a perplexity of 118.01. 

All three baseline models achieve reasonable perplexities indicating them to be good proxies for standard language models. 

\begin{table*}[hbt!]
\small
\centering
    \begin{tabular}{rrrrrrrrr}
     \toprule

      & \multicolumn{3}{c}{\textbf{Fixed Context}}
          & \multicolumn{3}{c}{\textbf{Infinite Context} } &  \multicolumn{1}{c}{}  \\
    \textbf{$\lambda$}  & \textbf{$\mu$} & \textbf{$\sigma$}  & \textbf{$\beta$} & \textbf{$\mu$} & \textbf{$\sigma$}  & \textbf{$\beta$} &   \textbf{$Ppl.$}\\
      \midrule      
     
train & 0.83 & 1.00 &  & 3.81 & 4.65 &  &  \\

0.0 & 0.74 & 0.91 & 0.40 & 2.23 & 2.90 & 0.38 & 62.56 \\

0.001 & 0.69 & 0.88 & 0.34 & 2.43 & 2.98 & 0.35 & 62.69 \\

\textbf{0.01} & \textbf{0.63} & \textbf{0.81} & \textbf{0.31} & 2.56 & 3.40 & 0.36 & 62.83 \\

\textbf{0.1} & 0.64 & 0.82 & 0.33 & \textbf{2.30} & \textbf{3.09} & \textbf{0.24} & \textbf{62.48} \\

0.5 & 0.70 & 0.91 & 0.39 & 2.91 & 3.76 & 0.38 & 62.5 \\

0.8 & 0.76 & 0.96 & 0.45 & 3.43 & 4.06 & 0.26 & 63.36 \\

1.0 & 0.84 & 0.94 & 0.38 & 2.42 & 3.02 & -0.30 & 62.63 \\
      


 \bottomrule
    \end{tabular}
    \caption{Experimental results for Penn Treebank and generated text for different $\lambda$  values}
     \label{tab:1}
\end{table*}
\begin{table*}[hbt!]
\small
\centering
    \begin{tabular}{rrrrrrrrr}
     \toprule
      & \multicolumn{3}{c}{\textbf{Fixed Context}}
          & \multicolumn{3}{c}{\textbf{Infinite Context} } & \multicolumn{1}{c}{}  \\
    \textbf{$\lambda$}  & \textbf{$\mu$} & \textbf{$\sigma$}  & \textbf{$\beta$} & \textbf{$\mu$} & \textbf{$\sigma$}  & \textbf{$\beta$} &   \textbf{$Ppl.$}\\
      \midrule

train & 0.80 & 1.00 &  & 3.70 & 4.60 &  &  \\

0.0 & 0.70 & 0.84 & 0.29 & 3.48 & 4.29 & 0.15 & \textbf{67.67} \\

0.001 & 0.69 & 0.84 & 0.27 & 2.32 & 3.12 & 0.16 & 67.84 \\

\textbf{0.01} & \textbf{0.61} & \textbf{0.79} & \textbf{0.20} & \textbf{1.88} & \textbf{2.69} & 0.14 & 67.78 \\

0.1 & 0.65 & 0.82 & 0.24 & 2.26 & 3.11 & \textbf{0.06} & 67.89 \\

0.5 & 0.70 & 0.88 & 0.31 & 2.25 & 3.17 & 0.20 & 69.07 \\

0.8 & 0.65 & 0.84 & 0.28 & 2.07 & 2.98 & 0.18 & 69.36 \\

1.0 & 0.74 & 0.92 & 0.27 & 2.32 & 3.21 & -0.08 & 69.56 \\
\bottomrule
\hline
    \end{tabular}
    \caption{Experimental results for  WikiText-2 and generated text for different $\lambda$  values}
    \label{tab:2}
\end{table*}
\begin{table*}[hbt!]
\small
\centering
    \begin{tabular}{rrrrrrrrr}
     \toprule
      & \multicolumn{3}{c}{\textbf{Fixed Context}}
          & \multicolumn{3}{c}{\textbf{Infinite Context} } & \multicolumn{1}{c}{}  \\
    \textbf{$\lambda$}  & \textbf{$\mu$} & \textbf{$\sigma$}  & \textbf{$\beta$} & \textbf{$\mu$} & \textbf{$\sigma$}  & \textbf{$\beta$} &   \textbf{$Ppl.$}\\
      \midrule

train & 0.72 & 0.94 &  & 0.77 & 1.05 &  &  \\

0.0 & 0.51 & 0.68 & 0.22 & 0.43 & 0.59 & 0.29 & 118.01 \\

0.1 & 0.38 & 0.52 & 0.19 & 0.85 & 1.38 & 0.22 & 116.49 \\

\textbf{0.5} & \textbf{0.34} & \textbf{0.48} & \textbf{0.14} & \textbf{0.79} & \textbf{1.31} & \textbf{0.20} & \textbf{116.19} \\

0.8 & 0.40 & 0.56 & 0.19 & 0.96 & 1.57 & 0.23 & 121.00 \\

1.0 & 0.62 & 0.83 & 0.21 & 1.71 & 2.65 & 0.31 & 120.55 \\

    
    \bottomrule
    \hline
    \end{tabular}
    \caption{Experimental results for CNN/Daily Mail and generated text for different $\lambda$  values}
    \label{tab:3}
\end{table*}

\subsection{Quantifying Biases} 
\label{bias}

 For numeric data,  bias can be caused simply by class imbalance, which is relatively easy to quantify and fix. For text and image data, the complexity in the nature of the data increases and it becomes difficult to quantify. Nonetheless, defining relevant metrics is crucial in assessing the bias exhibited in a dataset or in a model's behavior.

\subsubsection {Bias Score Definition}

In a text corpus, we can express the probability of a word occurring in context with gendered words as follows:

$$P(w|g)  =  \frac{{c(w,g)/ \Sigma_{i} c(w_i, g)}}{{c(g)}/\Sigma_{i} c(w_i)}\
$$ 
where $c(w,g)$ is a context window and $g$ is a set of gendered words that belongs to either of the two categories: male or female. For example, when $g = f$,  such words would include \textit{she}, \textit{her}, \textit{woman} etc. $w$ is any  word in the corpus, excluding stop words and gendered words. 
The bias score of a specific word $w$ is then defined as:
$$
bias_{train}(w) = \log\left({\frac{P(w|f)}{P(w|m)}}\right)
$$

This bias score is measured for each word in the text sampled from the training corpus and the text corpus generated by the language model. A positive bias score implies that a word cooccurs more often with female words than male words. For an infinite context, the words \textit{doctor} and \textit{nurse} would cooccur as many times with a female gender as with male gender words and the bias scores for these words will be equal to zero.

We conduct two sets of experiments where we define context window  $c(w,g)$ as follows:
\paragraph{Fixed Context} In this scenario, we take a fixed context window size and measure the bias scores. We generated bias scores for several context window sizes in the range $(5,15)$. For a context size $\textit{k}$, there are $\textit{k}$ words before and $\textit{k}$ words after the target word $w$ for which the bias score is being measured. Qualitatively, a smaller context window size has more focused information about the target word. On the other hand, a larger window size captures topicality \cite{levy2014dependency}. By choosing an optimal window of $\textit{k}=10$, we give equal weight of 5\% to the ten words before and the ten words after the target word.

\paragraph{Infinite Context} In this scenario, we take an infinite window of context with weights diminishing exponentially based on the distance between the target word $w$ and the gendered word $g$. This method emphasizes on the fact that the nearest word has more information about the target word. The farther the context gets away from a word, the less information it has about the word. We give 5\% weight to the words adjacent to the target word as in Fixed Context but reduce the weights of the words following by 5\% and 95\% to the rest; this applied recursively gives a base of 0.95. This method of exponential weighting instead of equal weighting adds to the stability of the measure.

\subsubsection{Bias Reduction Measures}
To evaluate debiasing of each model, we measure the bias for the generated corpus. 

$$bias_{\lambda}(w) = \log({\frac{P(w|f)}{P(w|m)})}
$$

To estimate the amplification or reduction of the bias, we fit a univariate linear regression model over bias scores of context words $w$ as follows:
$$
bias_{\lambda}(w)  = {\beta}*bias_{train}(w)+c
$$
where $\beta$ is the scaled amplification  measure relative to the training data. Reducing $\beta$ implies debiasing the model.

We also look at the distribution of the bias by evaluating mean absolute bias and deviation in bias scores for each context word in each of the generated corpora. 
$$
\mu_{\lambda} = mean(abs({bias_{\lambda}}))
; \sigma_{\lambda} = stdev({bias_{\lambda}})
$$
We take the mean of absolute bias score as the word can be biased in either of the two directions. 

\begin{table*}[t]
\small
\centering
    \begin{tabularx}{\linewidth}{lrX}
     \toprule
      
      \textbf{Target Word} &
    \textbf{$\lambda$}  & \textbf{Sample From Generated Text }\\
    \midrule
 \hfil   & 0.0 &``she was put on her own machine to raise money for her own wedding
\verb+<unk>+ route which saw her \textbf{crying} and \verb+<unk>+ down a programme today . effects began by bottom of her
marrow the \verb+<unk>+"\\
     crying& 0.5 & ``he \verb+<unk>+ in the americas with the \verb+<unk>+
which can spread a \verb+<unk>+ circumcision ceremony made last month . as he \verb+<unk>+ his mother s \verb+<unk>+ \textbf{crying} to
those that"
\\
     & 1.0 &``he discovered peaceful facebook remains when he was caught \textbf{crying} officers but was arrested after they found the
crash hire a man \verb+<unk>+ brown shocked his brother \verb+<unk>+ over"\\
\midrule

      & 0.0 & ``camilla said she talked to anyone and had previously left her love of two young children . it all comes with
her family in conviction of her son s death . it s been \textbf{fragile} . the \verb+<unk>+ and retail boy
that was rik s same maker identified genuinely \verb+<unk>+ attacked all''\\
      fragile & 0.5 & ``his children at nearby children s hospital in \verb+<unk>+ and went \verb+<unk>+ years after he was arrested on \verb+<unk>+ bail
. she spent relaxed weeks in prison after being sharply in \textbf{fragile} \verb+<unk>+ while she was jailed and strangled when
she was born in \verb+<unk>+ virginia''\\
      & 1.0 & ``could they possibly have a big barrier to jeff \verb+<unk>+ and \verb+<unk>+ my son all intelligence period that will
contain the east country s world from all in the world the truth is when we moved clear before the
split twenty days earlier that day . none of the distributed packs on the website can never \verb+<unk>+ re able
to \verb+<unk>+ it the second time so that fitting \textbf{fragile} \verb+<unk>+ are and less the country is \verb+<unk>+ . it
came as it was once \verb+<unk>+ million lead jobs mail yorkshire . adoption of these first product is ohio but
it is currently almost impossible for the moon to address and fully offshore hotly ''\\
    \midrule
      & 0.0 & ``mr \verb+<unk>+ worked traditions at the squadron base in \verb+<unk>+ rbs to marry the us government .he referring to the mainland them in february \verb+<unk>+ he kept communist \textbf{leadership} from undergoing"
\\
    leadership & 0.5 & ``obama s first wife janet had a chance to run the opposition for a superbowl event for charity the majority of the south african people s travel stage \verb+<unk>+ \textbf{leadership} while it was married off christmas"
\\
     & 1.0 &  ``the woman s lungs and drinking the ryder of his daughters s \textbf{leadership} morris said businesses . however being  of his mouth around wiltshire and burn talks from the hickey s \verb+<unk>+ employees"\\
      \midrule
      & 0.0 & 

``his legs and allegedly killed himself by suspicious points . in
the latest case after an online page he left \textbf{prisoner} in his home in \verb+<unk>+ near \verb+<unk>+ manhattan on
saturday when he was struck in his car operating in \verb+<unk>+ bay smoking \verb+<unk>+ and \verb+<unk>+ \verb+<unk>+ when he had"\\
      prisoner & 0.5 & ``it is something that the medicines can target \textbf{prisoner} and destroy \verb+<unk>+ firms in the
uk but i hope that there are something into the on top getting older people who have more branded them
as poor ."\\
      & 1.0 & ``the ankle follows a worker \verb+<unk>+
her \verb+<unk>+ \textbf{prisoner} she died this year before now an profile which clear her eye borrowed for her organ own
role . it was a huge accident after the drugs she had'' \\
    \bottomrule
    \end{tabularx}
    \caption{Generated text comparison for CNN/Daily Mail for different $\lambda$  values}
    \label{tab:4}
\end{table*}

\subsection{Model Debiasing}
Machine learning techniques that capture patterns in data to make coherent predictions can unintentionally capture or even amplify the bias in data \citep{Zhao2017men}. We consider a {\it gender subspace} present in the learned embedding matrix in our model as introduced in the \citet{bolukbasi2016man} paper. We train these embeddings on the word level language model instead of using the debiased pretrained embeddings \citep{Joel2019}. We conduct experiments for the three cases where we debias---input embeddings, output embeddings, and both the embeddings simultaneously.

Let $\mathbf{w} \in S_W$ be a word embedding corresponding to a word in the word embedding matrix $W$. Let $$D_i,\ldots, D_n \subset S_W$$ be the {\it defining sets\footnote{See the supplement for corpus-wise defining sets 
    }} that contain gender-opposing words, e.g. man and woman.
    The defining sets are designed separately for each corpus since certain words may not appear in another corpus. We consider it a defining set if both gender-opposing words occur in the training corpus. 
    
    If $\mathbf{u}_i$, $\mathbf{v}_i$  are the embeddings corresponding to the words \textit{man} and \textit{woman}, then $\{\mathbf{u}_i, \mathbf{v}_i\} = D_i$.
We consider the matrix $C$ which is defined as a stack of difference vectors between the pairs in the defining sets. We have:
\begin{align*}
C = 
\begin{bmatrix}
    (\frac{\mathbf{u}_1 - \mathbf{v}_1}{2})\\
    \vdots \\
    (\frac{\mathbf{u}_n - \mathbf{v}_n}{2})\\
\end{bmatrix}
= U \Sigma V
\end{align*}

The difference between the pairs encodes the gender information corresponding to the gender pair. We then perform singular value decomposition on $C$, obtaining $U \Sigma V$. The gender subspace $B$ is then defined as the first $k$ columns (where $k$ is chosen to capture $50\%$ of the variation) of the right singular matrix $V$: $$B = V_{1:k}$$

Let $N$ be the matrix consisting of the embeddings for which we would like the corresponding words to exhibit unbiased behavior. If we want the embeddings in $N$ to have minimal bias, then its projection onto the gender subspace $B$ should be small in terms its the squared Frobenius norm. Therefore, to reduce the bias learned by the embedding layer in the model, we can add the following bias regularization term to the training loss: 
$$ \mathcal{L}_B = \lambda \| N B \|_F^2$$
where $\lambda$ controls the importance of minimizing bias in the embedding matrix $W$ (from which $N$ and $B$ are derived) relative to the other components of the model loss. The matrices $N$ and $C$ are updated each iteration during the model training. 

We input 2000 random seeds in the language model as starting points to start word  generation. We use the previous words as an input to the language model and  perform multinomial selection  to generate up the next word. We repeat this up to 500 times. In total, we generate $10^{6}$ tokens for all three datasets for each $\lambda$ and measure the bias.

\section{Experiments}
\subsection{Model}
After achieving the baseline results, we run experiments to tune $\lambda$ as hyperparameter.  We report an in-depth analysis of bias measure on the models with debiased input embeddings.

\subsection{Results and Text Examples}We calculate the measures stated in Section \ref{bias} for the three datasets and the generated corpora using the corresponding RNN models. The results are shown in Tables \ref{tab:1}, \ref{tab:2} and \ref{tab:3}. We see that the $\mu$ consistently decline as we increase $\lambda$ until a point, beyond which the model becomes unstable. So there is a scope of optimizing the $\lambda$ values. The detailed analysis is presented in Section \ref{analysis}

Table \ref{tab:4}  shows excerpts around selected target words from the generated corpora to demonstrate the effect of debiasing for different values of $\lambda$.  We highlight the words \textit{crying} and \textit{fragile} that are typically associated with feminine qualities, along with the words \textit{leadership} and \textit{prisoners} that are stereotyped with male identity. These biases are reflected in the generated text for $\lambda=0$. We notice increased mention of the less probable gender in the subsequent generated text with debiasing ($\lambda=0.5, 1.0$). For \textit{fragile}, the generated text at $\lambda=1.0$ has reduced the mention of stereotyped female words but had no mentions of male words; resulting in a large chunk of neutral text. Similarly, in \textit{prisoners}, the generated text for $\lambda=0.5$ has no gender words. 

However, these are small snippets and the bias scores presented in the supplementary table quantifies the distribution of gender words around the target word in the entire corpus. These target words are chosen as they are commonly perceived gender biases and in our study, they show prominent debiasing effect.\footnote{For more examples, refer to the supplement}

\subsection{Analysis and Discussion}
\label{analysis}
 We consider a text corpus to be biased when it has a skewed distribution of words cooccuring with one gender vs another. Any dataset that has such demographic bias can lead to (potentially unintended) social exclusion  \cite{hovy2015demographic}. PTB and WikiText-2 consist of news articles  related to business, science, politics, and sports. These are all male dominated fields. However, CNN/Daily Mail consists of articles across diverse set of categories like entertainment, health, travel etc. Among the three corpora, Penn Treebank has more frequent mentions of male words with respect to female words and CNN/Daily Mail has the least. 
 
As defined, bias score of zero implies perfectly neutral word, any value higher/lower implies female/male bias. Therefore, the absolute value of bias score signifies presence of bias. Overall bias in a dataset can be estimated as the average of absolute bias score ($\mu$). The aggregated absolute bias scores $\mu$ of the three datasets---Penn Treebank,  WikiText-2, and CNN/Daily Mail---are 0.83, 0.80, and 0.72 respectively. Higher $\mu$ value in this measure means on-an-average the words in the entire corpus are more gender biased. As per the Tables \ref{tab:1}, \ref{tab:2}, and \ref{tab:3}, we see that the $\mu$ consistently decline as we increase $\lambda$ until a point, beyond which the model becomes unstable. So there is a scope of optimizing the $\lambda$ values. 

The second measure we evaluated is the standard deviation ($\sigma$) of the bias score distribution. Less biased dataset should have the bias score concentrating closer to zero and hence lower $\sigma$ value. We consistently see that, with the initial increase of $\lambda$, there is a decrease in $\sigma$ of the bias score distribution.

The final measure to evaluate debiasing is comparison of bias scores at individual word level. We regress the bias scores of the words in generated text against  their bias scores in the training corpus after removing the outliers. The slope of regression  $\beta$ signifies the amplification or dampening effect of the model relative to the training corpus. Unlike the previous measures, this measure gives clarity at word level bias changes. A drop in $\beta$ signifies reduction in bias and vice versa. A negative $\beta$ signifies inversion in bias assuming there are no other effects of the loss term. In our experiments, we observe $\beta$ to increase with higher values of $\lambda$ possibly due to instability in model and none of those values go beyond 1. 

We observe that corpus level bias scores like $\mu$, $\sigma$ are less effective measures to study efficacy of debiasing techniques because they fail to track the improvements at word level. Instead, we recommend a word level score comparison like $\beta$ to evaluate robustness of debiasing at corpus level.

To choose the context window in a more robust manner,  we take exponential weightings to the cooccurrences. The results for aggregated average of absolute bias and standard deviation show the same pattern as in fixed context window. 

As shown in the results above, we see that the standard deviation ($\sigma$), absolute mean ($\mu$) and slope of regression ($\beta$) reduce for smaller $\lambda$ relative to those in training data and then increase with $\lambda$ to match the variance in the original corpus. This holds for the experiments conducted with  fixed context window as well as with exponential weightings.

\section{Conclusion}
In this paper, we quantify and reduce gender bias in word level language models by defining a gender subspace and penalizing the projection of the word embeddings onto that gender subspace. We device a metric to measure gender bias in the training and the generated corpus.


In this study, we quantify corpus level bias in two different metrics---absolute mean ($\mu$) and standard deviation ($\sigma$). However, for evaluating debiasing effects, we propose a relative metric ($\beta$) to study the change in bias scores at word level in generated text vs. training corpus. To calculate $\beta$, we conduct an in-depth regression analysis of the word level bias measures in the generated text corpus over the same for the training corpus.

Although we found mixed results on amplification of bias as stated by \citet{Zhao2017men}, the debiasing method shown by \citet{bolukbasi2016man} was validated with the use of novel and robust bias measure designed in this paper. Our proposed methodology  can deal with distribution of words in a vocabulary in word level language model and it targets one way to measure bias, but it's highly likely that there is significant bias in the debiased models and data, just not bias that we can detect on this measure. It can be concluded different bias metrics  show different kinds of bias \citep{Gonen2019}.

We additionally observe a perplexity bias trade-off as a result of the additional bias regularization term. In order to reduce bias, there is a compromise on perplexity. Intuitively, as we reduce bias the perplexity is bound to increase due to the fact that, in an unbiased model, male and female words will be predicted with an equal probability.

\section{Acknowledgements}
We are grateful to Yu Wang and Jason Cramer for helping to initiate this project, to Nishant Subramani for helpful discussion, and to our reviewers for their thoughtful feedback. Bowman acknowledges support from Samsung Research.

\bibliography{naaclhlt2019.bbl}

\begin{thebibliography}{19}
\expandafter\ifx\csname natexlab\endcsname\relax\def\natexlab#1{#1}\fi

\bibitem[{Bolukbasi et~al.(2016)Bolukbasi, Chang, Zou, Saligrama, and
  Kalai}]{bolukbasi2016man}
Tolga Bolukbasi, Kai-Wei Chang, James~Y Zou, Venkatesh Saligrama, and Adam~T
  Kalai. 2016.
\newblock \href
  {http://papers.nips.cc/paper/6228-man-is-to-computer-programmer-as-woman-is-to-homemaker-debiasing-word-embeddings.pdf}
  {Man is to computer programmer as woman is to homemaker? {D}ebiasing word
  embeddings}.
\newblock In D.~D. Lee, M.~Sugiyama, U.~V. Luxburg, I.~Guyon, and R.~Garnett,
  editors, \emph{Advances in Neural Information Processing Systems 29}, pages
  4349--4357. Curran Associates, Inc.

\bibitem[{Caliskan et~al.(2017)Caliskan, Bryson, and Narayanan}]{caliskan2016}
Aylin Caliskan, Joanna~J. Bryson, and Arvind Narayanan. 2017.
\newblock \href {https://doi.org/10.1126/science.aal4230} {Semantics derived
  automatically from language corpora contain human-like biases}.
\newblock \emph{Science}, 356(6334):183--186.

\bibitem[{Font and Costa{-}Juss{\`{a}}(2019)}]{Joel2019}
Joel~Escud{\'{e}} Font and Marta~R. Costa{-}Juss{\`{a}}. 2019.
\newblock \href {http://arxiv.org/abs/1901.03116} {Equalizing gender biases in
  neural machine translation with word embeddings techniques}.
\newblock \emph{CoRR}, abs/1901.03116.

\bibitem[{Gonen and Goldberg(2019)}]{Gonen2019}
Hila Gonen and Yoav Goldberg. 2019.
\newblock Lipstick on a pig: Debiasing methods cover up systematic gender
  biases in word embeddings but do not remove them.
\newblock In \emph{Proceedings of the 2019 Conference of the North American
  Chapter of the Association for Computational Linguistics: Human Language
  Technologies}. Association for Computational Linguistics.

\bibitem[{Hermann et~al.(2015)Hermann, Kocisky, Grefenstette, Espeholt, Kay,
  Suleyman, and Blunsom}]{nips15_hermann}
Karl~Moritz Hermann, Tomas Kocisky, Edward Grefenstette, Lasse Espeholt, Will
  Kay, Mustafa Suleyman, and Phil Blunsom. 2015.
\newblock \href
  {http://papers.nips.cc/paper/5945-teaching-machines-to-read-and-comprehend.pdf}
  {Teaching machines to read and comprehend}.
\newblock In C.~Cortes, N.~D. Lawrence, D.~D. Lee, M.~Sugiyama, and R.~Garnett,
  editors, \emph{Advances in Neural Information Processing Systems 28}, pages
  1693--1701. Curran Associates, Inc.

\bibitem[{Hovy(2015)}]{hovy2015demographic}
Dirk Hovy. 2015.
\newblock Demographic factors improve classification performance.
\newblock In \emph{Proceedings of the 53rd Annual Meeting of the Association
  for Computational Linguistics and the 7th International Joint Conference on
  Natural Language Processing (Volume 1: Long Papers)}, volume~1, pages
  752--762.

\bibitem[{Lambrecht and Tucker(2018)}]{lambrecht2018algorithmic}
Anja Lambrecht and Catherine~E Tucker. 2018.
\newblock Algorithmic bias? an empirical study into apparent gender-based
  discrimination in the display of stem career ads.
\newblock \emph{Social Science Research Network (SSRN)}.

\bibitem[{Levy and Goldberg(2014)}]{levy2014dependency}
Omer Levy and Yoav Goldberg. 2014.
\newblock Dependency-based word embeddings.
\newblock In \emph{Proceedings of the 52nd Annual Meeting of the Association
  for Computational Linguistics (Volume 2: Short Papers)}, volume~2, pages
  302--308.

\bibitem[{Marcus et~al.(1993)Marcus, Santorini, and
  Marcinkiewicz}]{DBLP:journals/coling/MarcusSM94}
Mitchell~P. Marcus, Beatrice Santorini, and Mary~Ann Marcinkiewicz. 1993.
\newblock Building a large annotated corpus of english: The penn treebank.
\newblock \emph{Computational Linguistics}, 19(2):313--330.

\bibitem[{May et~al.(2019)May, Wang, Bordia, Bowman, and
  Rudinger}]{Chandler2019}
Chandler May, Alex Wang, Shikha Bordia, Samuel~R. Bowman, and Rachel Rudinger.
  2019.
\newblock On measuring social bias in sentence encoders.
\newblock In \emph{Proceedings of the 2019 Conference of the North American
  Chapter of the Association for Computational Linguistics: Human Language
  Technologies}. Association for Computational Linguistics.

\bibitem[{Merity et~al.(2018)Merity, Keskar, and Socher}]{merity2017}
Stephen Merity, Nitish~Shirish Keskar, and Richard Socher. 2018.
\newblock \href {https://openreview.net/forum?id=SyyGPP0TZ} {Regularizing and
  optimizing {LSTM} language models}.
\newblock In \emph{International Conference on Learning Representations}.

\bibitem[{Merity et~al.(2016)Merity, Xiong, Bradbury, and
  Socher}]{merity2016pointer}
Stephen Merity, Caiming Xiong, James Bradbury, and Richard Socher. 2016.
\newblock Pointer sentinel mixture models.
\newblock \emph{arXiv preprint arXiv:1609.07843}.

\bibitem[{Mikolov et~al.(2010)Mikolov, Karafi{\'{a}}t, Burget, Cernock{\'{y}},
  and Khudanpur}]{mikolov2010}
Tomas Mikolov, Martin Karafi{\'{a}}t, Luk{\'{a}}s Burget, Jan Cernock{\'{y}},
  and Sanjeev Khudanpur. 2010.
\newblock Recurrent neural network based language model.
\newblock In \emph{{INTERSPEECH}}, pages 1045--1048. {ISCA}.

\bibitem[{Mikolov et~al.(2013)Mikolov, Sutskever, Chen, Corrado, and
  Dean}]{mikolov2013distributed}
Tomas Mikolov, Ilya Sutskever, Kai Chen, Greg~S Corrado, and Jeff Dean. 2013.
\newblock Distributed representations of words and phrases and their
  compositionality.
\newblock In \emph{Advances in neural information processing systems}, pages
  3111--3119.

\bibitem[{Ngan and Grother(2015)}]{ngan2015face}
Mei Ngan and Patrick Grother. 2015.
\newblock \emph{Face recognition vendor test (FRVT) performance of automated
  gender classification algorithms}.
\newblock US Department of Commerce, National Institute of Standards and
  Technology.

\bibitem[{Recasens et~al.(2013)Recasens, Danescu-Niculescu-Mizil, and
  Jurafsky}]{recasens2013linguistic}
Marta Recasens, Cristian Danescu-Niculescu-Mizil, and Dan Jurafsky. 2013.
\newblock Linguistic models for analyzing and detecting biased language.
\newblock In \emph{Proceedings of the 51st Annual Meeting of the Association
  for Computational Linguistics (Volume 1: Long Papers)}, volume~1, pages
  1650--1659.

\bibitem[{Rudinger et~al.(2018)Rudinger, Naradowsky, Leonard, and {Van
  Durme}}]{rudinger-EtAl:2018:N18}
Rachel Rudinger, Jason Naradowsky, Brian Leonard, and Benjamin {Van Durme}.
  2018.
\newblock Gender bias in coreference resolution.
\newblock In \emph{Proceedings of the 2018 Conference of the North American
  Chapter of the Association for Computational Linguistics: Human Language
  Technologies}, New Orleans, Louisiana. Association for Computational
  Linguistics.

\bibitem[{Zhao et~al.(2017)Zhao, Wang, Yatskar, Ordonez, and
  Chang}]{Zhao2017men}
Jieyu Zhao, Tianlu Wang, Mark Yatskar, Vicente Ordonez, and Kai{-}Wei Chang.
  2017.
\newblock Men also like shopping: Reducing gender bias amplification using
  corpus-level constraints.
\newblock In \emph{{EMNLP}}, pages 2979--2989. Association for Computational
  Linguistics.

\bibitem[{Zhao et~al.(2018)Zhao, Zhou, Li, Wang, and Chang}]{zhao2018b}
Jieyu Zhao, Yichao Zhou, Zeyu Li, Wei Wang, and Kai-Wei Chang. 2018.
\newblock \href {http://aclweb.org/anthology/D18-1521} {Learning gender-neutral
  word embeddings}.
\newblock In \emph{Proceedings of the 2018 Conference on Empirical Methods in
  Natural Language Processing}, pages 4847--4853. Association for Computational
  Linguistics.

\end{thebibliography}
\bibliographystyle{acl_natbib}


\appendix 

\section{Defining sets}
The gender pair list for each corpus is designed separately. We consider only those gender pairs that occur in the training corpus. Below are the gender lists corresponding to each corpus: 
\subsection{Penn Treebank} 
\textbf{Male Words:} 
``actor"
``boy"
``father"
``he"
``him "
``his"
``male"
``man"
``men"
``son"
``sons"
``spokesman"
``wife"
``king"
``brother"

\noindent\textbf{Female Words:} 
``actress"
``girl "
``mother"
``she"
``her "
``her"
``female"
``woman"
``women"
``daughter"
``daughters"
``spokeswoman"
``husband"
``queen"
``sister"
\subsection{WikiText-2}

\textbf{Male Words:} ``actor"
``Actor"
``boy"
``Boy"
``boyfriend"
``Boys"
``boys"
``father"
``Father"
``Fathers"
``fathers"
``Gentleman"
``gentleman"
``gentlemen"
``Gentlemen"
``grandson"
``he"
``He"
``hero"
``him"
``Him"
``his"
``His"
``Husband"
``husbands"
``King"
``kings"
``Kings"
``male"
``Male"
``males"
``Males"
``man"
``Man"
``men"
``Men"
``Mr."
``Prince"
``prince"
``son"
``sons"
``spokesman"
``stepfather"
``uncle"
``wife"
``king"

\noindent\textbf{Female Words:} 
``actress"
``Actress"
``girl"
``Girl"
``girlfriend"
``Girls"
``girls"
``mother"
``Mother"
``Mothers"
``mothers"
``Lady"
``lady"
``ladies"
``Ladies"
``granddaughter"
``she"
``She"
``heroine"
``her"
``Her"
``her"
``Her"
``Wife"
``wives"
``Queen"
``queens"
``Queens"
``female"
``Female"
``females"
``Females"
``woman"
``Woman"
``women"
``Women"
``Mrs."
``Princess"
``princess"
``daughter"
``daughters"
``spokeswoman"
``stepmother"
``aunt"
``husband"
``queen"
\subsection{CNN/Daily Mail}

\textbf{Male Words:} ``actor"
``boy"
``boyfriend"
``boys"
``father"
``fathers"
``gentleman"
``gentlemen"
``grandson"
``he"
``him"
``his"
``husbands"
``kings"
``male"
``males"
``man"
``men"
``prince"
``son"
``sons"
``spokesman"
``stepfather"
``uncle"
``wife"
``king"
``brother"
``brothers"

\noindent\textbf{Female Words:} 
``actress"
``girl"
``girlfriend"
``girls"
``mother"
``mothers"
``lady"
``ladies"
``granddaughter"
``she"
``her"
``her"
``wives"
``queens"
``female"
``females"
``woman"
``women"
``princess"
``daughter"
``daughters"
``spokeswoman"
``stepmother"
``aunt"
``husband"
``queen"
``sister"
``sisters"

\section{Word Level Bias Examples}
 Tables \ref{tab:wm} and \ref{tab:wf} show the bias scores at individual word level for selected words for Wikitext-2. The tables show how the scores vary for the training text and the generated text for different values of  $\lambda$

\noindent Tables \ref{tab:dmm} and \ref{tab:dmf} show the bias scores at individual word level for selected words for CNN/Daily Mail. The tables show how the scores vary for the training text and the generated text for different values of  $\lambda$

\begin{table*}[b]
\small
\centering
    \begin{tabular}{|l|r|r|r|r|r|r|r|}
     \toprule
     
    \textbf{Target Words}  & \textbf{$training$} & \textbf{$\lambda$}{=0.0}  &
    \textbf{$\lambda$}{=0.01}&
    \textbf{$\lambda$}{=0.1}& \textbf{$\lambda$}{=0.5} & \textbf{$\lambda$}{=0.8} & \textbf{$\lambda$}{=1.0}   \\
      \midrule

Arts       & -0.76 & -1.20 & -0.87 & -0.32 & -0.17 & 0.13  & -1.48 \\
Boston     & -0.95 & -1.06 & -0.23 & -1.06 & -0.13 & -0.37 & -0.94 \\
Edward     & -0.68 & -1.06 & 0.09  & -0.56 & -0.14 & -0.44 & -0.23 \\
George     & -0.52 & -0.91 & -0.26 & -0.22 & -0.48 & -0.26 & 0.01  \\
Henry      & -0.59 & -1.06 & 0.11  & -0.34 & -0.84 & -0.92 & -0.61 \\
Peter      & -0.69 & -2.06 & -0.09 & -0.14 & -0.32 & 0.08  & 0.53  \\
Royal      & -0.01 & -1.89 & -0.39 & -0.61 & -0.64 & -1.14 & -0.56 \\
Sir        & -0.01 & -1.76 & -0.99 & -0.86 & -0.64 & -0.16 & 0.07  \\
Stephen    & -0.35 & -1.20 & -0.18 & -1.01 & -0.84 & -0.11 & 0.36  \\
Taylor     & -0.84 & -0.91 & 0.57  & 0.00  & -0.01 & -0.39 & -0.83 \\
ambassador & -0.76 & -1.20 & -0.23 & -0.63 & -0.74 & 0.43  & -0.81 \\
failed     & -0.46 & -2.06 & 0.03  & -0.36 & -1.00 & 0.17  &       \\
focused    & -0.22 & -0.91 & -0.12 & -0.41 & -0.40 & -0.57 & -0.53 \\
idea       & -0.20 & -1.06 & -0.36 & -0.16 & -0.27 & -0.06 & -0.42 \\
manager    & -1.58 & -1.60 & -0.04 & -0.30 & -1.08 & -0.30 & -1.06 \\
students   & -0.60 & -0.79 & -0.31 & -0.29 & -0.32 & -0.51 & -0.50 \\
university & -0.12 & -1.06 & 0.17  & -1.01 & -0.79 & -0.95 & -0.70 \\
wife       & -0.92 & -1.29 & -0.81 & -1.02 & -0.57 & -0.67 & -1.03 \\
work       & -0.24 & -0.88 & -0.48 & -0.23 & -0.49 & -0.52 & -0.13 \\
youth      & -0.39 & -1.20 & 0.54  & -0.16 & -0.68 & 0.58  &      
\\
\bottomrule
    \end{tabular}
    \caption{WikiText-2 bias scores for the words biased towards male gender for  different  $\lambda$  values}
    \label{tab:wm}
\end{table*}

\begin{table*}[t]
\small
\centering
    \begin{tabular}{|l|l|r|r|r|r|r|r|r|}
     \toprule
     
    \textbf{Target Words}  & \textbf{$training$} & \textbf{$\lambda$}{=0.0}  & 
      \textbf{$\lambda$}{=0.01}  &
    \textbf{$\lambda$}{=0.1}& \textbf{$\lambda$}{=0.5} & \textbf{$\lambda$}{=0.8} & \textbf{$\lambda$}{=1.0}   \\
      \midrule

Katherine    & 1.78 & 2.27 & 1.38  & 0.69 & 0.95 & 0.75  & 0.70  \\
Zenobia      & 0.05 & 0.88 & 1.84  & 0.47 & 0.65 & 1.24  &       \\
childhood    & 0.48 & 1.80 & 0.12  & 1.10 & 0.37 & 0.38  & 0.34  \\
cousin       & 0.13 & 0.88 & 0.67  & 0.13 & 0.09 & 0.67  & 0.71  \\
humor        & 0.34 & 1.29 &       & 0.69 & 0.61 & 0.34  &       \\
invitation   & 0.19 & 1.80 & -0.87 & 0.69 & 0.57 & -0.44 & -0.25 \\
parents      & 0.51 & 0.76 & 0.77  & 0.08 & 0.45 & 0.57  & 1.11  \\
partners     & 0.85 & 2.27 & -0.28 & 0.98 & 0.87 & -0.17 & 3.22  \\
performances & 0.79 & 1.02 & -0.20 & 0.16 & 0.03 & 0.10  & -1.80 \\
producers    & 1.04 & 1.58 & 0.33  & 0.78 & 1.35 & -1.45 & 0.18  \\
readers      & 0.22 & 0.88 & 0.28  & 0.29 & 0.36 & -0.32 & -1.29 \\
stars        & 0.85 & 1.58 & 0.16  & 0.90 & 0.46 & -0.28 & -0.08 \\
talent       & 0.02 & 0.88 & -0.75 & 0.10 & 0.31 & -0.86 &       \\
wore         & 0.09 & 0.88 & 0.48  & 0.29 & 0.65 & 0.16  & -0.69\\

 \bottomrule
    \end{tabular}
    \caption{WikiText-2 bias scores for the words biased towards female gender for  different $\lambda$  values}
    \label{tab:wf}
\end{table*}

\begin{table*}[t]
\small
\centering
    \begin{tabular}{|l|r|r|r|r|r|r|}
     \toprule
     
    \textbf{Target Words}  & \textbf{$training$} & \textbf{$\lambda$}{=0.0}  & \textbf{$\lambda$}{=0.1}& \textbf{$\lambda$}{=0.5} & \textbf{$\lambda$}{=0.8} & \textbf{$\lambda$}{=1.0}   \\
      \midrule

abusers         & -0.66 & -1.17 & -0.56 & -0.77 & -0.16 & -1.93 \\
acting          & -0.23 & -0.81 & -0.59 & -0.35 & -0.54 & 0.60  \\
actions         & -0.27 & -0.51 & -0.06 & -0.07 & -0.53 & -0.45 \\
barrister       & -1.35 & -2.00 & -0.64 & -0.76 & -0.08 & -0.69 \\
battle          & -0.27 & -0.53 & -0.10 & -0.32 & -0.16 & 0.21  \\
beneficiary     & -1.64 & -1.87 & -1.06 & -0.22 & 0.63  &       \\
bills           & -0.32 & -0.53 & -0.18 & -0.50 & 0.23  & 0.69  \\
businessman     & -0.19 & -1.81 & -0.71 & -0.45 & -0.53 & -1.93 \\
cars            & -0.43 & -0.55 & -0.32 & -0.11 & -0.24 & -0.27 \\
citizen         & -0.03 & -0.30 & -0.03 & -0.22 & -0.01 & 0.04  \\
cocaine         & -0.59 & -1.00 & -0.84 & -0.44 & -0.42 & -0.32 \\
conspiracy      & -0.57 & -0.73 & -0.66 & -0.39 & -0.83 & -0.43 \\
controversial   & -0.21 & -0.39 & -0.39 & -0.02 & -0.17 & -0.43 \\
cooking         & -0.48 & -0.53 & -0.24 & -0.22 & 0.07  & -0.52 \\
cop             & -1.30 & -1.42 & -0.77 & -0.72 & 0.00  & 0.26  \\
drug            & -0.76 & -0.82 & -0.53 & -0.42 & -0.54 & -0.63 \\
executive       & -0.04 & -0.34 & -0.22 & -0.04 & -0.48 & -0.36 \\
fighter         & -0.59 & -0.90 & -0.48 & -0.36 & -0.89 & -0.11 \\
fraud           & -0.17 & -0.30 & -0.16 & -0.19 & 0.10  & -0.63 \\
friendly        & -0.48 & -0.53 & -0.30 & -0.23 & 0.36  & -0.21 \\
heroin          & -0.57 & -0.67 & -0.28 & -0.26 & -0.66 & 0.49  \\
journalists     & -0.25 & -1.08 & -0.55 & -0.76 & -0.44 &       \\
lawyer          & -0.39 & -0.47 & -0.14 & -0.10 & -0.20 & -0.50 \\
lead            & -0.47 & -0.50 & -0.40 & -0.09 & -0.07 & -0.32 \\
leadership      & -0.25 & -0.74 & -0.28 & -0.68 & -0.57 & -0.99 \\
notorious       & -0.18 & -0.64 & -0.36 & -0.22 & -0.12 & -1.49 \\
offensive       & -0.17 & -0.39 & -0.28 & -0.17 & -0.52 & -0.11 \\
officer         & -0.25 & -0.29 & -0.21 & -0.13 & -0.17 & -0.19 \\
outstanding     & -0.25 & -1.55 & -0.98 & -0.50 & 0.03  & 0.04  \\
parole          & -0.54 & -0.86 & 0.00  & -0.08 & 0.07  & -1.30 \\
pensioners      & -0.48 & -0.86 & -0.77 & -0.07 & 0.64  & 0.40  \\
prisoners       & -0.52 & -0.99 & -0.18 & -0.29 & -0.17 & -1.59 \\
religion        & -0.41 & -0.97 & -0.15 & -0.48 & 0.18  & -1.68 \\
reporters       & -0.60 & -0.93 & -0.26 & -0.05 & -0.52 & -0.97 \\
representatives & -0.07 & -0.48 & -0.40 & -0.18 & -0.46 & -0.83 \\
research        & -0.34 & -0.46 & -0.05 & -0.33 & 0.03  & -0.58 \\
resignation     & -0.95 & -1.67 & -0.61 & -0.58 & -0.40 & -1.12 \\
sacrifice       & -0.03 & -1.08 & -0.38 & -0.17 & -1.29 &       \\
supervisor      & -0.66 & -0.92 & -0.44 & -0.25 & -0.17 & 0.48  \\
violent         & -0.17 & -0.54 & -0.07 & -0.22 & -0.19 & -0.19\\
 \bottomrule
    \end{tabular}
    \caption{ CNN/Daily Mail bias scores for the words biased towards male gender for  different  $\lambda$  values}
    \label{tab:dmm}
\end{table*}

\begin{table*}[t]
\small
\centering
    \begin{tabular}{|l|r|r|r|r|r|r|}
     \toprule
     
    \textbf{Target Words}  & \textbf{$training$} & \textbf{$\lambda$}{=0.0}  & \textbf{$\lambda$}{=0.1}& \textbf{$\lambda$}{=0.5} & \textbf{$\lambda$}{=0.8} & \textbf{$\lambda$}{=1.0}   \\
      \midrule

abusive       & 0.00 & 0.40 & 0.06 & 0.39 & 0.48  & -0.65 \\
appealing     & 0.44 & 1.22 & 0.23 & 0.30 & -0.68 & 1.16  \\
bags          & 0.34 & 1.42 & 0.48 & 0.05 & 0.16  & 0.64  \\
beloved       & 0.17 & 0.35 & 0.27 & 0.15 & 0.52  & 0.36  \\
carol         & 0.76 & 1.41 & 0.20 & 0.39 & 0.27  & 0.48  \\
chatted       & 0.03 & 1.83 & 0.20 & 0.19 & -0.14 & -0.25 \\
children      & 0.29 & 0.46 & 0.36 & 0.26 & 0.41  & 0.27  \\
comments      & 0.17 & 0.46 & 0.04 & 0.02 & -0.35 & -0.14 \\
crying        & 0.28 & 0.70 & 0.19 & 0.57 & 0.17  & 0.87  \\
designer      & 0.73 & 0.80 & 0.57 & 0.69 & 0.53  & -1.53 \\
designers     & 0.44 & 2.14 & 1.29 & 0.76 & -0.11 & 1.11  \\
distressed    & 0.15 & 0.53 & 0.23 & 0.26 & -0.56 & 1.36  \\
divorced      & 0.68 & 0.70 & 0.18 & 0.10 & 0.31  & 0.88  \\
dollar        & 0.44 & 1.63 & 0.65 & 0.59 & -0.24 &       \\
donated       & 0.52 & 0.57 & 0.06 & 0.15 & 0.68  & 0.26  \\
donating      & 1.29 & 1.38 & 0.27 & 0.80 & -0.03 & -0.21 \\
embracing     & 1.13 & 1.78 & 0.74 & 0.55 & 1.48  & -0.94 \\
encouragement & 0.85 & 0.94 & 0.22 & 0.50 & 0.37  & 0.55  \\
endure        & 0.85 & 0.94 & 0.26 & 0.29 & 1.02  &       \\
expecting     & 1.01 & 1.07 & 0.26 & 0.12 & 0.53  & 0.06  \\
feeling       & 0.21 & 0.84 & 0.16 & 0.25 & 0.16  & 0.29  \\
festive       & 0.15 & 0.53 & 0.52 & 0.14 & 0.21  & 0.26  \\
fragile       & 0.44 & 0.94 & 0.20 & 0.45 & -0.20 &       \\
happy         & 0.32 & 0.66 & 0.10 & 0.11 & 0.11  & 0.25  \\
healthy       & 0.52 & 0.64 & 0.26 & 0.45 & 0.24  & 0.25  \\
hooked        & 0.78 & 1.38 & 0.12 & 0.12 & -0.11 & -0.09 \\
hurting       & 0.75 & 1.13 & 0.33 & 0.34 & 0.44  & 0.26  \\
indian        & 0.18 & 0.28 & 0.15 & 0.02 & -0.02 & -0.26 \\
kissed        & 0.31 & 1.03 & 0.17 & 0.19 & 0.28  & -0.22 \\
kissing       & 0.26 & 1.14 & 0.54 & 0.61 & 0.44  & -0.14 \\
loving        & 0.41 & 0.73 & 0.43 & 0.18 & 0.15  & -0.34 \\
luxurious     & 0.59 & 0.82 & 0.17 & 0.44 & -0.03 & -0.83 \\
makeup        & 1.60 & 1.63 & 0.07 & 0.22 & 1.09  &       \\
mannequin     & 0.95 & 1.92 & 0.70 & 0.04 & 1.42  &       \\
married       & 0.29 & 0.37 & 0.34 & 0.09 & 0.30  & 0.42  \\
models        & 0.35 & 1.22 & 0.28 & 0.38 & 0.90  & 0.08  \\
pictures      & 0.08 & 0.50 & 0.10 & 0.04 & -0.06 & 0.59  \\
pray          & 0.62 & 1.58 & 0.25 & 0.35 & -0.25 & 0.96  \\
relationship  & 0.53 & 0.62 & 0.39 & 0.32 & 0.58  & 0.43  \\
scholarship   & 0.80 & 1.16 & 0.80 & 0.70 & 0.53  & 0.45  \\
sharing       & 0.58 & 0.73 & 0.33 & 0.67 & 0.42  & 0.17  \\
sleeping      & 0.18 & 0.71 & 0.27 & 0.35 & 0.56  & 0.58  \\
stealing      & 0.10 & 0.48 & 0.32 & 0.18 & 0.06  & -0.53 \\
tears         & 0.50 & 0.58 & 0.44 & 0.12 & 0.45  & 0.35  \\
thanksgiving  & 0.85 & 2.14 & 1.14 & 1.08 & 0.90  &       \\
waist         & 1.33 & 1.45 & 0.68 & 0.02 & 0.31  & 0.96 \\
 \bottomrule
    \end{tabular}
    \caption{ CNN/Daily Mail bias scores for the words biased towards female gender for  different $\lambda$  values}
    \label{tab:dmf}
\end{table*}

\end{document}